This is a preprint copy that has been accepted for publication in *IEEE Transactions on Cybernetics*.

Please cite this article as:

Yukun Bao, Tao Xiong, Zhongyi Hu, "PSO-MISMO Modeling Strategy for Multi-Step-Ahead Time Series Prediction". *IEEE Transactions on Cybernetics*. 2013, accepted. doi:10.1109/TCYB.2013. 2265084.

# PSO-MISMO Modeling Strategy for Multi-Step-Ahead Time Series Prediction

Yukun Bao, Member, IEEE, Tao Xiong, Zhongyi Hu

*Abstract*—**Multi-step-ahead time series prediction is one of the most challenging research topics in the field of time series modeling and prediction, and is continually under research. Recently, the multiple-input several multiple-outputs (MISMO) modeling strategy has been proposed as a promising alternative for multi-step-ahead time series prediction, exhibiting advantages compared with the two currently dominating strategies, the iterated and the direct strategies. Built on the established MISMO strategy, this study proposes a particle swarm optimization (PSO)-based MISMO modeling strategy, which is capable of determining the number of sub-models in a self-adaptive mode, with varying prediction horizons. Rather than deriving crisp divides with equal-size *s* prediction horizons from the established MISMO, the proposed PSO-MISMO strategy, implemented with neural networks, employs a heuristic to create flexible divides with varying sizes of prediction horizons and to generate corresponding sub-models, providing considerable flexibility in model construction, which has been validated with simulated and real datasets.**

*Index Terms*—**Multi-step-ahead time series prediction, multiple-output models, particle swarm optimization, genetic algorithm.**

## I. INTRODUCTION

Despite intensive research efforts in time series modeling and prediction, most models continue to be limited to one-step-ahead prediction rather than multi-step-ahead prediction, primarily because of the challenges arising from increased uncertainty from longer prediction horizons. The most common modeling strategies for multi-step-ahead time series prediction rely either on iterated or direct strategies [1-3]. An iterated strategy first constructs a one-step-ahead prediction model, and then uses the predicted values as known data to predict the next ones. In the direct strategy, a set of $H$ prediction models are solved to predict values $H$ steps ahead. The accumulation of errors in the iterated case deteriorates the accuracy of the prediction, and the direct strategy requires a sharply increased computational expense, especially in the case of a longer prediction horizon. Based upon these two mainstream modeling strategies, there has been a revival of interest in developing novel multi-step-ahead prediction strategies such as the DirRec strategy [4], the multiple-input multiple-output (MIMO) strategy [5] and multiple-input several multiple-outputs (MISMO) strategy [6, 7] (DIRMO in [8]). A review and comparative study on strategies for multi-step-ahead time series has indicated that the multiple-output strategies, including MIMO and MISMO (DIRMO), are the best performing approaches, which has been validated with NN5 forecasting competition data [8].

This study focuses on a MISMO strategy, improving it in terms of modeling flexibility and more accurate prediction. MISMO is achieved by transforming the original $H$-step-ahead prediction task into $n = H/s$ subtasks, yielding $n$ sub-models with a fixed number of multiple outputs, $s$, for each, where $1 \leq s \leq H$. The value of $s$ is chosen through cross-validation, analyzing the performance of the MISMO model for different values of $s$ on the learning set, and then using the best value to estimate the outputs [6, 7]. In the special case where $s = H$, MISMO can be regarded as MIMO, and when $s = 1$, MISMO is the same as a direct strategy. The superior out-of-sample performance of the MISMO strategy presented in [7] depends on the trade-off between heterogeneity and computational complexity. The MISMO strategy exploits the heterogeneity across subtasks efficiently, with less computational complexity than the direct prediction technique. An important feature of the MISMO strategy is that the original prediction modeling task is divided into several subtasks with an equal number of outputs (prediction horizons). Handling indivisibility in the problem is an explicit difficulty with the method. In addition, an underlying problem arises from the crisp divides on the prediction horizons with equal size $s$ that induce a latent separation of dependencies among the steps in the different sub models. This problem may prevent the MISMO strategy from considering complex dependencies between prediction steps within different sub models and consequently reduce the prediction accuracy. Thus a strategy with more flexibility and a mechanism to self-adaptively determine the prediction horizon divides while modeling is appealing. This study proposes an improved PSO-MISMO

Manuscript received September 12, 2012. This work was supported in part by the Natural Science Foundation of China under Grant No. 70771042, the Fundamental Research Funds for the Central Universities (2012QN208-HUST), the MOE (Ministry of Education in China) Project of Humanities and Social Science (Project No.13YJA630002), and a grant from the Modern Information Management Research Center at Huazhong University of Science and Technology.

Yukun Bao (corresponding author), is with school of management, Huazhong University of Science and Technology, Wuhan, P.R.China (Tel: 86-27-87558579; fax: 86-27-87556437; e-mail: yukunbao@hust.edu.cn).

Tao Xiong, is with school of management, Huazhong University of Science and Technology, Wuhan, P.R.China (e-mail: taoxiong@hust.edu.cn).

Zhongyi Hu, is with school of management, Huazhong University of Science and Technology, Wuhan, P.R.China (e-mail: huzhyi21@hust.edu.cn).

modeling strategy for multi-step-ahead prediction that incorporates a heuristic based on particle swarm optimization (PSO) into the MISMO modeling process to self-adaptively determine the number of sub-models with varying prediction horizons.

For the purpose of justification, this study compares the rank of the proposed modeling strategy with the four well-established strategies (i.e., iterated, direct, MIMO and MISMO) with neural networks (NNs). In addition, as a straightforward alternative, binary genetic algorithm-based MISMO modeling strategy (GA-MISMO) is selected as counterpart against the proposed PSO-MISMO and compared as well. Both simulated (i.e., Logistic and Mackey-Glass time series) and real (i.e., NN3 competition data) datasets are used for the comparisons.

The paper is structured as follows. Section II provides a brief review of the multiple-output modeling strategies, including MIMO and MISMO, with limited details. Section III describes the basic concepts of FNN for multi-step-ahead prediction and the implementation of the FNN-based prediction model in the current study. The concept of the binary PSO is explained in Section IV. In Section V, details of the proposed PSO-MISMO strategy are presented. Section VI details the experimental setting on data, accuracy measure, genetic algorithm as alternative, and experimental procedure. Experimental results are discussed in Section VII, and concluding remarks are provided in Section VIII.

## II. MIMO AND MISMO STRATEGIES

Multi-step-ahead prediction can be described as an estimation of $\varphi_{N+h}, (h = 1, 2, \ldots, H)$, where $H$ is an integer greater than one, given the current and previous observation $\varphi_t, (t = 1, 2, \ldots, N)$. In this section, two competing modeling strategies, MIMO and MISMO, with multiple-input multiple-output structures for multi-step-ahead prediction are described, to enable understanding of the proposed PSO-MISMO strategy in Section V.

### A. MIMO Strategy

MIMO was first proposed in [5] and characterized as a multiple-input multiple-output approach, where the predicted value is not a scalar quantity but a vector of future values $(\varphi_{N+1}, \varphi_{N+2}, \ldots, \varphi_{N+H})$ of the time series $\varphi_t, (t = 1, 2, \ldots, N)$. Compared with the direct strategy, which estimates $\varphi_{N+h}, (h = 1, 2, \ldots, H)$ using $H$ models, MIMO employs only one multiple-output model, preserving the temporal stochastic dependency hidden in the predicted time series.

The MIMO modeling strategy learns one multiple-output forecasting model as follows:

$$(\varphi_{i+1}, \ldots, \varphi_{i+H}) = f(\varphi_i, \ldots, \varphi_{i-d+1}) + \varepsilon \qquad (1)$$

where $d$ is the maximum embedding order and $\varepsilon$ is a vector noise term of zero mean and nondiagonal covariance.

After the learning process, the estimations of the $H$ next values are returned according to the following equation:

$$(\widehat{\varphi}_{i+1}, \ldots, \varphi_{i+H}) = \hat{f}(\varphi_i, \ldots, \varphi_{i-d+1}) \qquad (2)$$

The limitation of MIMO lies in forcing all the prediction horizons to be predicted with the same model structure (for instance, the same inputs $(\varphi_i, \ldots, \varphi_{i-d+1})$ ), resulting in a reduction of the modeling effort, but likely to bias the returned model results.

### B. MISMO Strategy

A solution to the shortcomings of MIMO was proposed in [6, 7], where the constraint of the fixed modeling structure is relaxed by tuning an integer parameter $s$, which calibrates the dimensionality of the outputs on the basis of a validation criterion. For a given $s$, MISMO generates $n = H/s$ sub-models, which may have different inputs from each other but a fixed size of outputs at $s$, resulting in slightly more modeling flexibility.

The MISMO modeling strategy learns $n$ multiple-output sub-models as follows:

$$(\varphi_{i+(k-1)s}, \ldots, \varphi_{i+ks-1}) = f_k(\varphi_i, \ldots, \varphi_{i-d+1}) + \varepsilon, k \in \{1, \ldots, n\}. \quad (3)$$

where $d$ is the maximum embedding order and $\varepsilon$ is a zero-mean vector of size $s$.

After the learning process, the estimations of the $H$ next values are returned by the following:

$$(\widehat{\varphi}_{i+(k-1)s}, \ldots, \varphi_{i+ks-1}) = \hat{f}_k(\varphi_i, \ldots, \varphi_{i-d+1}), k \in \{1, \ldots, n\}. \qquad (4)$$

The MISMO approach highlights the trade-off between the property of preserving the stochastic dependency among the predicted values and the flexibility of the modeling procedure [8]. However, the increased flexibility of the MISMO strategy requires an additional parameter s, which gives rise to several issues in itself. An explicit issue is handling indivisibility when computing $n = H/s$ , where all the parameters should be integers. Furthermore, the number of outputs of every sub-model derived from MISMO is fixed at $s$ . This induces a latent separation of dependencies among the steps that may prevent the MISMO strategy from considering complex dependencies between prediction steps within different sub-models and consequently reduce the prediction accuracy. Further study has been solicited to determine the divides of prediction horizons adaptively while modeling rather than fixing them at $s$ .

## III. FEED-FORWARD NEURAL NETWORKS

Although many types of neural networks (NNs) have been proposed, the most popular one for time series prediction is the feed-forward NNs (FNNs). Because the focus of this study is the evaluation of the proposed modeling strategy relative to the other competitors for multi-step-ahead prediction, a simple three-layer FNN is implemented to model the selected strategies, capable of creating a common and reliable benchmark. In addition, the proposed modeling strategy and the four well-established strategies can all be implemented in the FNN because of the flexible structure of the FNN. Thus, the standard three-layer FNN with nodes in adjacent layers fully connected is used in this study. In this section, a short introduction to feed-forward neural networks for multi-step-ahead prediction is provided, and then the implementation of the FNN-based prediction model in the current study is described.

### A. FNN for Multi-Step-Ahead Prediction

Fig. 1 shows an example of a typical three-layer FNN model with four input nodes, four hidden nodes (one hidden layer), and two output neurons used for multi-step-ahead prediction. The input nodes are the previous lagged observations while the outputs provide the forecast for the future values in a multi-step-ahead fashion. Hidden nodes with appropriate nonlinear activation functions are used to process the information received by the input nodes. Bias (or intercept) terms are used both in the hidden and the output layers. To facilitate the understanding of the FNN for multi-step-ahead time series prediction, the FNN with the MIMO modeling strategy will be used as an example in this section.

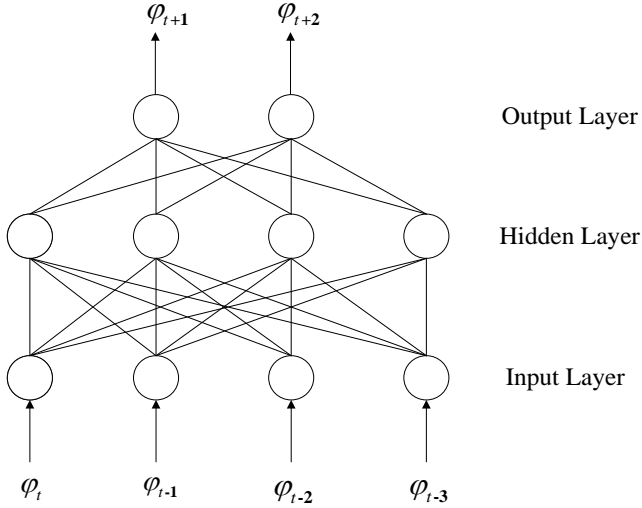

Fig. 1. A three-layer feed-forward neural network

Given the training sample $D = \left\{ (\mathbf{x}_i, \mathbf{y}_i) \in \left( \Re^d \times \Re^H \right) \right\}_{i=d}^N$, the FNN model for $H$-step-ahead prediction can be written as follows:

$$\mathbf{y} = \left( \sum_{j=1}^k w_{jh}^0 \sigma \left( \sum_{i=1}^n w_{ij}^r x_i + b_j^r \right) \right) + b_h^0, \ h = 1, \ldots, H \quad (5)$$

where $\mathbf{x}$ is the input vector $(\varphi_i, \ldots, \varphi_{i-d+1})$, $\mathbf{y}$ is the output vector $(\varphi_{i+1}, \ldots, \varphi_{i+H})$, $\mathbf{w}^r$, $\mathbf{w}^0$, $\mathbf{b}^r$, and $\mathbf{b}^0$ are the weights and biases of hidden and output layers, $n$ and $H$ are the sizes of input and output vectors, $k$ is the number of hidden units, and $\sigma(\cdot)$ is a nonlinear activation function.

Functionally, equation (5) can also be written as follows:

$$(\varphi_{i+1}, \ldots, \varphi_{i+H}) = \boldsymbol{\vartheta}(\varphi_i, \ldots, \varphi_{i-d+1}, \ ) + \boldsymbol{\varepsilon} \quad (6)$$

where $f(\cdot)$ represents the FNN model and $\boldsymbol{\theta}$ is a vector that contains all the parameters in equation (5). The determination of parameter set $\boldsymbol{\theta}$ is based on the local error between the measured $(\varphi_{i+1}, \ldots, \varphi_{i+H})$ and predicted values $(\hat{\varphi}_{i+1}, \ldots, \varphi_{i+H})$. In the current study, the mean square error (MSE) is used as the index of multi-step-ahead prediction performance to be minimized, defined as follows:

$$\text{MSE} = \sum_{h=1}^H \left( \hat{\varphi}_{i+h} - \varphi_{i+h} \right)^2 \bigg/ H \quad (7)$$

### B. Implementation of the FNN-based Prediction Model

Following the procedure from [9], the design tasks for the FNN-based prediction model can be roughly divided into four parts: data preprocessing, FNN designing, FNN implementation and validation. The implementation of the FNN-based model for multi-step-ahead prediction is presented step-by-step below:

- Data preprocessing. Normalization is a standard requirement for time series modeling and prediction. Thus, the data sets are first scaled by linear transference to map onto a range of [0, 1]. After the linear transference, deseasonalization and detrending are performed when necessary. Deseasonalization is performed with the most recent X-12-ARIMA seasonal adjustment procedure. Detrending is performed by fitting a polynomial time trend to the data, and then subtracting the estimated trend from the series when trends are detected by the Mann-Kendall test.
- FNN designing. Selecting an appropriate architecture for the FNN is normally the first step when designing an FNN-based prediction system. The standard three-layer FNN with fully connected nodes in adjacent layers is used in this study. The number of input nodes, hidden nodes, output nodes, and the type of activation functions are defined in the present study as follows:

1) Selecting the number of input nodes: The number of input nodes is determined by the input selection. The filter method, which selects a set of inputs by optimizing a criterion over different combinations of inputs by means of a search algorithm, is employed for input selection in the current study. The filter method requires the setting of two elements: the criterion, i.e., the statistic which estimates the quality of the selected variables, and the search algorithm, which describes the policy used to explore the input space. With respect to the criterion, the partial mutual information [10] is used for the models using the iterated or direct strategies, and an extension of the Delta test [6] is used for the models with MIMO, MISMO, GA-MISMO, or PSO-MISMO strategies. With respect to the search algorithm, a forward-backward selection method that offers the flexibility to reconsider input variables previously discarded and to discard input variables previously selected is used. The maximum embedding order, $d$, is set to 15.

2) Selecting the number of hidden nodes: The number of the hidden nodes cannot be determined in advance. Thus empirical experimentations are needed to determine this number [11]. Because of the small sample size of many of the series, the experimentation is limited to five possible values of hidden nodes, i.e., 2, 4, 6, 8, or 10. The best number of hidden nodes is determined by using the original Akaike's information criterion (AIC).

3) Selecting the number of output nodes: The number of output nodes is determined by the modeling strategy. Most of the modeling strategies found in the literature can be classified into two categories, single-output structure and multiple-output structure. In the single-output structure strategies, such as the iterated and direct strategies, multiple inputs map to a single output. The MIMO, MISMO, GA-MISMO, and PSO-MISMO strategies are all based on the multiple-output structure that maps multiple inputs to multiple outputs. Therefore, for the iterated and direct strategies, only one output node is used for the FNN model. When employing the MIMO strategy, the number of output nodes is equal to the number of prediction horizons. In the MISMO strategy, the original $H$-step-ahead prediction tasks are separated into $n = H/s$ subtasks, resulting in $n$ sub-models with fixed multiple outputs, $s$, for each, where $1 \le s \le H$. Unlike the MISMO strategy, which uses a fixed number of outputs $s$ for each sub-model, the GA-MISMO and PSO-MISMO strategies self-adaptively determines the number of sub-models with varying number of outputs, as described in Section V.

4) Selecting the type of activation functions: A sigmoid transfer function is used at each hidden node and a linear transfer function is used at each output node.

• FNN implementation. The Levenberg and Marquardt algorithm (LMA) provided by the MATLAB (Version 2009b) NN toolkit is used for the training. Node biases are applied at the hidden and output layers. For the stopping criterion, the number of learning epochs is chosen as 1000, having no prior knowledge of the appropriate value. To determine the optimal parameters for the FNN, the common practice of fivefold cross-validation is used.

• Validation: For each series of the simulated and real dataset, the last 18 observations are separated for *ex-ante* performance assessment (out-of-sample), statistically independent from the parameter estimation process. The remainder of the data (in-sample) is used for parameter estimation and the FNN modeling.

## IV. Binary Particle Swarm Optimization

Particle swarm optimization (PSO) [12] is a population-based self-adaptive search algorithm that exploits a population of individuals to probe promising regions of the search space. In the PSO, the population is referred to as a swarm that consists of a number of particles. Each particle represents a potential solution of the optimization task and has a position represented by a vector. Each particle moves to a new position according to both local information (the particle memory) and global information (the knowledge of all the particles). Thus, the PSO has the ability to converge to local and/or global optimal solutions in a small number of generations. In this section, a brief introduction to a variant of the original PSO, i.e., binary PSO, is presented, to understand the mechanism of the proposed PSO-MISMO strategy applied in this study.

The PSO was initially designed to solve continuous optimization problems. Since the introduction of PSO, successful applications to several optimization problems [13-17] have demonstrated its potential. However, the major obstacle of applying a PSO successfully is its continuous nature. The first variant of the PSO algorithm for solving problems with binary-value solution elements (binary PSO) [18] was also developed by the creator of the original PSO. There are two key differences between the binary PSO and the original PSO. The first difference is the representation of the particle. In the binary PSO, every particle is characterized by a binary solution representation (Note that the representation is primarily related to a coding issue in the proposed PSO-MISMO strategy). The other difference is that the velocity of a particle in the binary PSO is a probability vector, where each probability element determines the likelihood of that binary variable having the value of one. Fig. 2 lists the pseudocode algorithm for the basic binary PSO.

---

**Algorithm 1 Binary particle swarm optimization**

**Initialize** a population of particles with random positions and velocities, throughout the input space.

**While** not sufficiently good performance or maximum number of iterations **do**

    **for** each particle $i$ *in* $\{1, \ldots, I\}$ **do**

        **if** $f(P_i) < f(pbest_i)$ **then**

            $pbest_i = P_i$

        **end if**

    **end**

    $pbest_{\min} =$

        $\left\{ pbest_{i*} \middle| f(pbest_{i*}) = \min\left(f(pbest_i), i \in \{1, \ldots, I\}\right) \right\}$

    **if** $f(pbest_{\min}) < f(gbest)$ **then**

        $gbest = pbest_{\min}$

    **end if**

    **for** each particle $i$ *in* $\{1, \ldots, I\}$ **do**

        Update velocity $V_i$ according to equation (8).

        Update position $P_i$ according to equation (11).

    **end**

**end while**

---

Fig. 2. Pseudocode for the binary PSO algorithm

The binary PSO updates the velocity according to equation (8) as follows:

$$V_i^t = w \cdot V_i^{t-1} + c_1 \cdot rand1 \cdot \left(pbest_i^{t-1} - P_i^{t-1}\right)$$
$$+ c_2 \cdot rand2 \cdot \left(gbest^{t-1} - P_i^{t-1}\right) \quad (8)$$

where $c_1$ and $c_2$ are the cognitive and interaction coefficients, $rand1$ and $rand2$ are random real numbers uniformly distributed between 0 and 1. $w$, the inertia weight, a user-specified parameter controls the momentum of the particle. A larger inertia weight pushes towards global exploration while a smaller inertia weight helps in fine-tuning the current search area. The following weighting function is usually utilized:

$$w = (w_{max} - w_{min}) \times \frac{T - t}{t} + w_{min} \quad (9)$$

where $w_{max}$ and $w_{min}$ are initial weight and final weight, respectively, $T$ is the maximum number of allowable iterations and $t$ is the current iteration.

The velocity is constrained to the interval [0, 1] by using the following sigmoid transformation:

$$S(V_i^t) = \frac{1}{1 - \exp(-V_i^t)} \quad (10)$$

where $S\left(V_i^t\right)$, is the probability that bit $V_i^t$ is equal to 1. To avoid $S\left(V_i^t\right)$ approaching 0 or 1, a constant $V_{max}$ is often set at 4 such that $v_i^t \in [-4, 4]$, then the following relationships are applied:

$$V_i^t = \begin{cases} V_{max} & if \ V_i^t > V_{max} \\ -V_{max} & if \ V_i^t > -V_{max} \\ V_i^t & otherwise \end{cases} \tag{11}$$

Each particle, at each time step, changes its current position according to equation (10), based on equation (12), as follows:

$$P_i^t = \begin{cases} 1 & if \ rand3 < S\left(V_i\left(t\right)\right) \\ 0 & otherwise \end{cases} \tag{12}$$

where $rand3$ is a random real number uniformly distributed between 0 and 1.

## V. PROPOSED PSO-MISMO MODELING STRATEGY

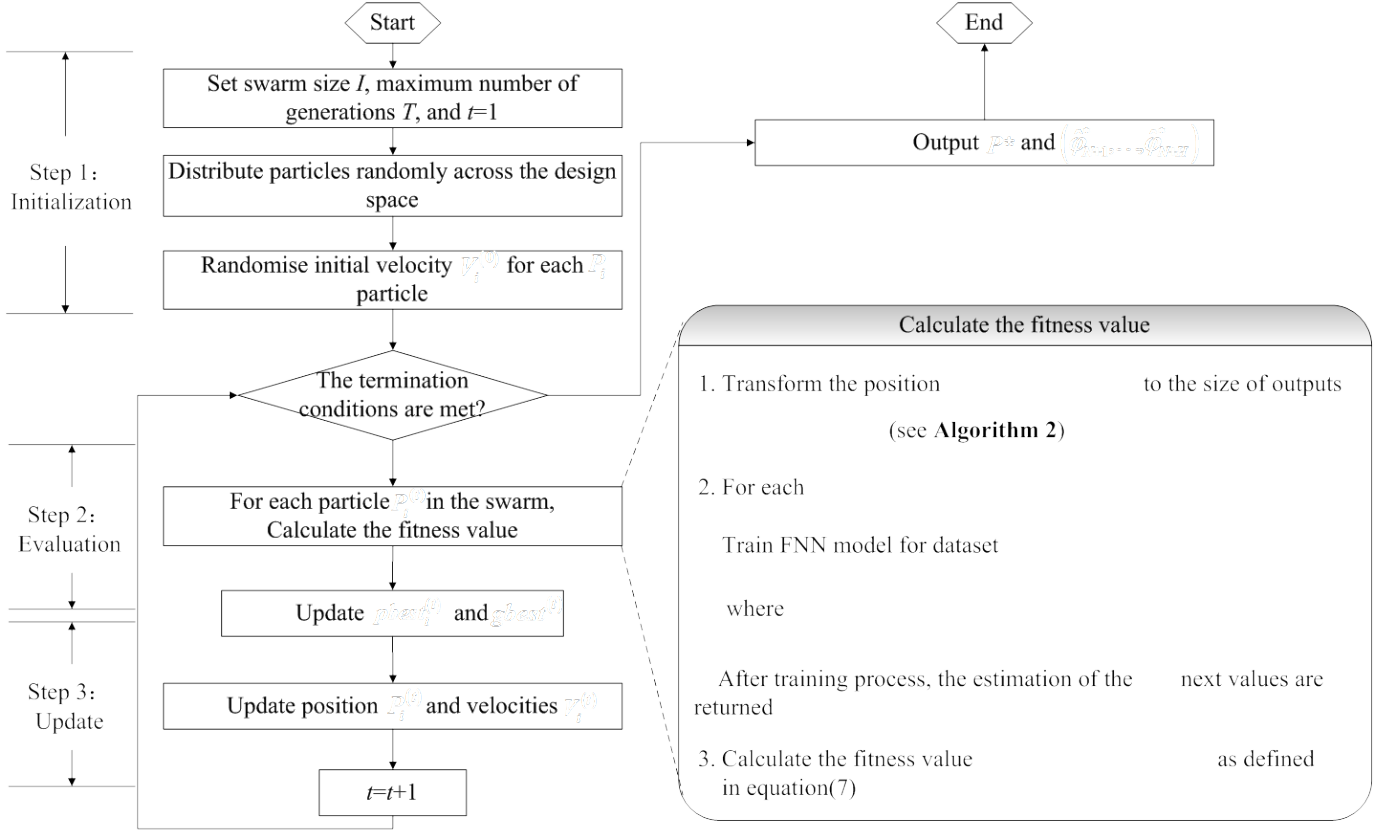

Fig. 3. Flowchart of the proposed PSO-MISMO modeling strategy

To address the limitations of MISMO mentioned first in Section I and illustrated further in Section II.B, an improved binary PSO-based MISMO is developed that self-adaptively determines the number of sub-models with varying prediction horizons. The proposed strategy is abbreviated PSO-MISMO.

Fig. 3 depicts the flowchart of the proposed PSO-MISMO modeling strategy. As shown in Fig. 3, the detailed PSO-MISMO algorithm consists of three major operations: initialization, evaluation, and update. Before describing details of these operations, coding issues will be addressed. The overall learning process in Fig. 3 is elaborated step by step below.

- Coding. Consider a time series $\{\varphi_1, \ldots, \varphi_N\}$ composed of $N$ observations for which the next $H$ observations are to be predicted. To avoid the restriction associated with adopting a fixed $s$ in the original MISMO method, a flexible vector $\mathbf{S} = (s_1, \ldots, s_J)$ is used, where $J$ is the number of prediction tasks, and $s_j$ represents the number of consecutive outputs to be predicted at a time by the $\text{FNN}_j$ model (Note that $\text{FNN}_j$ model refers to the $j^{th}$ sub model with $s_j$ outputs). The values of $J$ and $\mathbf{S} = (s_1, \ldots, s_J)$ are determined by the binary vector $\mathbf{P} = (p_1, \ldots, p_{H-1})$ that represents the segmentation point (see **Algorithm 2**). Each particle $\mathbf{P} = (p_1, \ldots, p_{H-1})$, representing a possible solution to the optimization problem, is determined by the binary PSO algorithm. The values of $H-1$ components of the binary vector $\mathbf{P} = (p_1, \ldots, p_{H-1})$ are either 0 or 1; if the value of a variable is 0, then the original task will not be divided by this segmentation point. If the value of a variable is 1, then the prediction task will be divided by this segmentation point. Fig. 4 illustrates the solution representation.

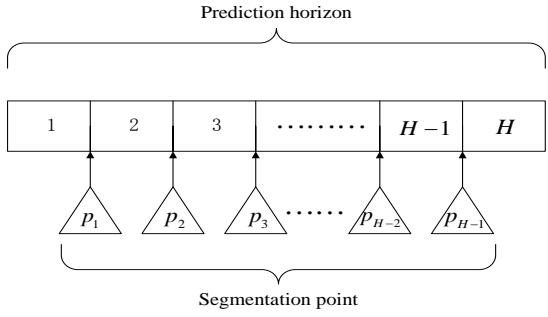

Fig. 4. Solution representation

In the special case that $\{p_i = 1, i = 1, 2, \ldots, H-1\}$, the strategy equals the direct strategy, and when $\{p_i = 0, i = 1, 2, \ldots, H-1\}$, it equals the MIMO strategy.

For further illustration, assume a 10-steps-ahead prediction task ($H = 10$). Given the binary vector $\mathbf{P} = (0, 0, 1, 0, 0, 0, 0, 1, 0)$, then $\mathbf{S} = (3, 5, 2)$ and $J = 3$ (see **Algorithm 2**). Thus, the original task is separated into three prediction subtasks as follows:

$$(\varphi_{i+1}, \varphi_{i+2}, \varphi_{i+3}) = f_1(\varphi_i, \varphi_{i-1}, \ldots, \varphi_{i-d+1}) + \boldsymbol{\varepsilon_1}$$
$$(\varphi_{i+4}, \varphi_{i+5}, \varphi_{i+6}, \varphi_{i+7}, \varphi_{i+8}) = f_2(\varphi_i, \varphi_{i-1}, \ldots, \varphi_{i-d+1}) + \boldsymbol{\varepsilon_2}$$
$$(\varphi_{i+9}, \varphi_{i+10}) = f_3(\varphi_i, \varphi_{i-1}, \ldots, \varphi_{i-d+1}) + \boldsymbol{\varepsilon_3}$$

where $\{\boldsymbol{\varepsilon_i}, i = 1, 2, 3\}$ is a zero-mean vector of size 3, 5, 2.

After the learning process, the estimation of the ten next values is returned by the three steps as follows:

$$(\hat{\varphi}_{i+1}, \hat{\varphi}_{i+2}, \hat{\varphi}_{i+3}) = \hat{f}_1(\varphi_i, \varphi_{i-1}, \ldots, \varphi_{i-d+1})$$
$$(\hat{\varphi}_{i+4}, \hat{\varphi}_{i+5}, \hat{\varphi}_{i+6}, \hat{\varphi}_{i+7}, \hat{\varphi}_{i+8}) = \hat{f}_2(\varphi_i, \varphi_{i-1}, \ldots, \varphi_{i-d+1})$$
$$(\hat{\varphi}_{i+9}, \hat{\varphi}_{i+10}) = \hat{f}_3(\varphi_i, \varphi_{i-1}, \ldots, \varphi_{i-d+1})$$

The prediction horizons of these prediction sub-models (or the number of outputs of the three FNN models) are 3, 5, and 2, respectively. This difference, or heterogeneity, is the highlight of the proposed strategy. For clarity, the flow of this example is illustrated in Fig. 5.

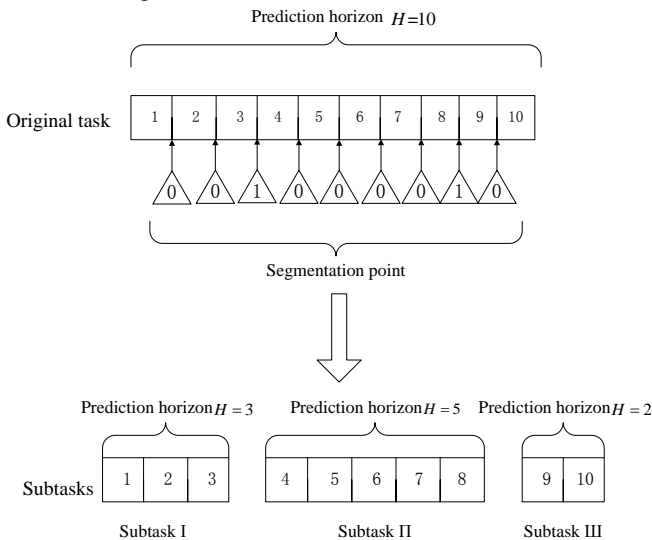

Fig. 5. An example of solution representation

**Algorithm 2.** Computing the value of $\mathbf{S} = (s_1, \ldots, s_J)$

**Input**: $H$, prediction horizon.

**Input**: $\mathbf{P} = (p_1, \ldots, p_{H-1})$, binary vector.

**Output**: $\mathbf{S} = (s_1, \ldots, s_J)$, $s_j$ represents the number of consecutive outputs to be predicted at a time by the FNN$_j$ model

$J = \sum_{i=1}^{H-1} p_i + 1$

$N = \sum_{i=1}^{H-1} p_i$

 $SP$ is the vector of size $N$

 $n = 1$

**for** $i$ in $\{1, \ldots, H-1\}$ **do**

  **if** $p_i == 1$ **then**

  $SP[n] = i$

  $n = n + 1$

  **end if**

**end**

$s_1 = SP[1]$

$s_J = H - SP[N]$

**for** $j$ in $\{2, \ldots, J-1\}$ **do**

  $s_j = SP[j] - SP[j-1]$

**end**

**Return** $\mathbf{S} = (s_1, \ldots, s_J)$

Fig. 6. Pseudocode for computing the value of $\mathbf{S} = (s_1, \ldots, s_J)$

• Initialization. All the particles are randomly generated with 0 or 1 using a 50% probability. All the velocity components are assigned the initial value of 0. Concerning the selection of parameters (i.e., inertial weight $w$, coefficients $c_1$ and $c_2$, swarm size $I$, and number of iterations $T$) in binary PSO, it is yet another challenging model selection task [19]. Fortunately, several empirical and theoretical studies have been performed about the parameters of PSO from which valuable information can be obtained [20-23]. In this study, the parameters are determined according to the recommendations in these studies and selected based on the prediction performance and computational time in a trial-error fashion. Table I summaries the final parameters.

TABLE I
PARAMETER SELECTION OF BINARY PSO

| Parameters | Values |
|---|---|
| Swarm size $I$ | 20 |
| Number of iterations $T$ | 100 |
| Cognitive coefficient $c_1$ | 2.0 |
| Interaction coefficient $c_2$ | 2.0 |
| Initial weight $w_{max}$ | 0.9 |
| Final weight $w_{min}$ | 0.4 |

• Evaluation. In the current generation $t$, for each particle $p_i^{(t)}$, the position $\mathbf{P} = (p_1, \ldots, p_{H-1})$ is transformed to $\mathbf{S} = (s_1, \ldots, s_J)$ (see **Algorithm 2**). Then, each $s_j \in \{s_1, \ldots, s_J\}$ derived from the particle $p_i^{(t)}$, is trained using the training FNN$_j$ model and the dataset $D_j = \left\{ (\mathbf{x}_j, \mathbf{y}_j) \in (\Re^m \times \Re^{s_j}) \right\}_{i=d}^N$, where: $\mathbf{x}_j \subset \{\varphi_{n-d+1}, \ldots, \varphi_n\}$, $\mathbf{y}_j = \left\{ \varphi_{n+1+\sum_{k=1}^{j} s_{(k-1)}}, \ldots, \varphi_{n+\sum_{k=1}^{j} s_k} \right\}$, $s_0 = 0$.

After the training process, the estimations of the next values of $s_j$ are returned. Once all $H$ next values for particle $p_i^{(t)} \in \{p_1^{(t)}, \ldots, p_I^{(t)}\}$ are obtained, the fitness value of each particle $p_i^{(t)}$ is calculated according to $\text{MSE}_i^t = \sum_{h=1}^{H} \left( \hat{\varphi}_{N+h}^i - \varphi_{N+h}^i \right)^2 \Big/ H$, where $\hat{\varphi}_{N+h}^i$ is the $h$-step-ahead forecast for particle $p_i^{(t)}$, $\varphi_{N+h}^i$ is the true value for particle $p_i^{(t)}$, and $H$ is the prediction horizon.

• Update. In the current generation, once the fitness values of all the particles in the swarm are calculated, the *pbest* of each particle, and the *gbest* of the swarm are obtained. Then, all the particles' velocities and positions are updated according to equations (8) and (12), respectively. If the termination conditions are met, then output *gbest* (that is, $\mathbf{P^*}$) and the corresponding $H$-step-ahead forecast $\{\hat{\varphi}_{N+1}^*, \ldots, \hat{\varphi}_{N+H}^*\}$ for particle $\mathbf{P^*}$; otherwise, go back to the previous step.

## VI. EXPERIMENTAL SETTING

### A. Datasets Description

To evaluate the performances of the proposed PSO-MISMO and the counterparts in terms of the forecast accuracy, two simulated time series, i.e., Logistic and Mackey-Glass time series, and a real world dataset, i.e., NN3 competition dataset, are used in this present study.

Logistic and Mackey-Glass time series are recognized as benchmark time series that have been commonly used and reported by a number of studies related to time series modeling and forecasting [24-26].

The Logistic map is a demographic model that was popularized by May [27] as an example of a simple non-linear system that exhibits complex, chaotic behavior. It is drawn from equation (13).

$$\varphi_i = 4\varphi_{i-1}\left(1 - \varphi_{i-1}\right) \quad (13)$$

The Mackey-Glass time series is approximated from the differential equation (14) (see [28]).

$$\frac{d\varphi_i}{di} = \frac{0.2\varphi_{i-\tau}}{1 + \varphi_{i-\tau}^{10}} - 0.1\varphi_i \quad (14)$$

For each data-generating process (DGP), i.e., Logistic and Mackey-Glass process, we simulate twenty time series with different initialization and sample size, as was shown in Table II. The data for these time series are generated by the *Chaotic*

*Systems Toolbox*[1] from the MATLAB software.

TABLE II
INITIALIZATION AND SAMPLE SIZE OF THE SIMULATED TIME SERIES

| No. | DGP | | | Sample size |
|---|---|---|---|---|
| | Logistic ( $\varphi_i$ ) | Mackey-Glass ( $\varphi_i$ , $\tau$ ) | | |
| 1 | 0.100 | 1.000 | 15 | 485 |
| 2 | 0.125 | 1.200 | 15 | 496 |
| 3 | 0.150 | 1.400 | 15 | 523 |
| 4 | 0.175 | 1.600 | 15 | 548 |
| 5 | 0.200 | 1.800 | 15 | 674 |
| 6 | 0.225 | 2.000 | 15 | 692 |
| 7 | 0.260 | 1.000 | 16 | 726 |
| 8 | 0.275 | 1.200 | 16 | 758 |
| 9 | 0.300 | 1.400 | 16 | 779 |
| 10 | 0.325 | 1.600 | 16 | 791 |
| 11 | 0.350 | 1.800 | 16 | 821 |
| 12 | 0.375 | 2.000 | 16 | 843 |
| 13 | 0.400 | 1.000 | 17 | 869 |
| 14 | 0.425 | 1.200 | 17 | 889 |
| 15 | 0.450 | 1.400 | 17 | 912 |
| 16 | 0.475 | 1.600 | 17 | 926 |
| 17 | 0.510 | 1.800 | 17 | 946 |
| 18 | 0.525 | 2.000 | 17 | 964 |
| 19 | 0.550 | 1.000 | 18 | 987 |
| 20 | 0.575 | 1.200 | 18 | 1002 |

The NN3 competition was organized in 2007, targeting computational-intelligence forecasting approaches. The competition dataset of 111 monthly time series drawn from homogeneous population of real business time series is used for evaluation[2].

As such, three datasets of 20 Logistic time series, 20 Mackey-Glass time series, and 111 NN3 time series are used for evaluating the performances of the proposed PSO-MISMO and the counterparts in this study. Each series is split into an estimation sample and a hold-out sample. The last 18 observations are saved for evaluating and comparing the out-of-sample forecast performances of the various multi-step-ahead prediction strategies. All performance comparisons are based on these $18 \times 20$ out-of-sample points for Logistic and Mackey-Glass datasets and $18 \times 111$ out-of-sample points for NN3 datasets.

### B. Accuracy Measure

For each prediction horizon $h$, three alternative forecast accuracy measures are considered: the mean absolute percentage error (MAPE), the symmetric mean absolute percentage error (SMAPE), and the mean absolute scaled error (MASE). The definitions of each are as follows:

$$\text{MAPE}_h = \frac{1}{M} \sum_{m=1}^{M} \left| \frac{\varphi_{N+h}^m - \hat{\varphi}_{N+h}^m}{\varphi_{N+h}^m} \right| * 100 \quad (15)$$

$$\text{SMAPE}_h = \frac{1}{M} \sum_{m=1}^{M} \left| \frac{\varphi_{N+h}^m - \hat{\varphi}_{N+h}^m}{\varphi_{N+h}^m + \hat{\varphi}_{N+h}^m} \right| * 100 \quad (16)$$

---

$$\text{MASE}_h = \frac{1}{M} \sum_{m=1}^{M} \left| \frac{\varphi_{N+h}^m - \hat{\varphi}_{N+h}^m}{\frac{1}{N-1} \sum_{i=2}^{N} |\varphi_i^m - \varphi_{i-1}^m|} \right| \qquad (17)$$

where $\hat{\varphi}_{N+h}^m$ is the $h$- step-ahead forecast for time series $m$, $\varphi_{N+h}^m$ is the true time series value for series $m$, $H$ is the prediction horizon (in this case, $H = 18$) and $M$ is the number of time series in the datasets (in this case, $M = 20$ for Logistic and Mackey-Glass datasets and $M = 111$ for NN3 datasets). For preprocessed data, including normalization, deseasonalization and detrending, we convert the outputs from the neural networks back to their original scales. Performance prediction can then be compared directly.

### C. Genetic Algorithm as Alternative

To further assess the performance of the proposed PSO-MISMO modeling strategy for multi-step-ahead prediction, we compare the experimental results of the PSO-MISMO with those produced by other discrete evolutionary algorithms, such as genetic algorithm (GA) (for detailed introduction to GA, please refer to [29-31]). This subsection briefly presents the implementation of GA-MISMO, a straightforward alternative against PSO-MISMO.

Fig. 7 depicts the flowchart of the GA-MISMO modeling strategy. As shown in Fig. 7, the detailed GA-MISMO algorithm consists of three major parts: initialization, evaluation, and operation. Before describing details of these parts, coding issues will be addressed. The overall learning process in Fig. 7 is elaborated step by step below.

• Coding. The chromosome design in GA-MISMO is the same with the particle design in PSO-MISMO, as was presented in Section V. Actually, the individual $\mathbf{P} = (p_1, \ldots, p_{H-1})$ may be regarded as chromosome in terms of GA-MISMO, or as particle in terms of PSO-MISMO. The values of $H - 1$ components of the binary vector $\mathbf{P} = (p_1, \ldots, p_{H-1})$ are either 0 or 1: if the value of a variable is 0, then the original task will not be divided by this segmentation point; if the value of a variable is 1, then the prediction task will be divided by this segmentation point. The coding issue in GA-MISMO is the same as PSO-MISMO and will not be presented in details here to save space.

TABLE III
PARAMETER SELECTION OF GA

| Parameters | Values |
|---|---|
| Population size | 20 |
| Number of iterations | 100 |
| Crossover probability | 0.90 |
| Mutation probability | 0.02 |

• Initialization. All the chromosomes are randomly generated with 0 or 1 using a 50% probability. In setting the parameters of GA, it is required to do extensive simulations to find suitable values for various parameters. In this study, GA's parameters are determined through preliminary simulation and selected according to the recommendations in [32-34] in a trial-error fashion. Table III summarizes the final parameter of GA.

• Evaluation. The evaluation in GA-MISMO is the same as

PSO-MISMO, as was presented in Section V.

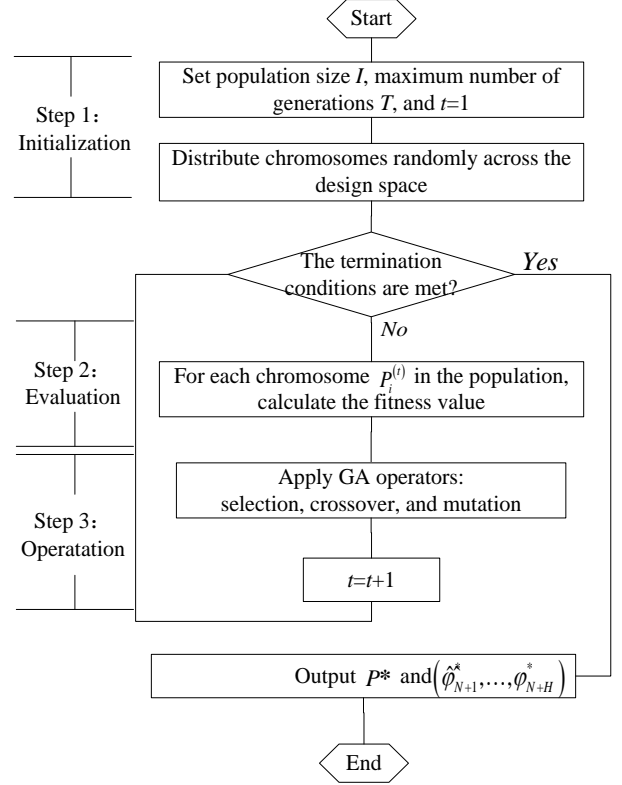

Fig. 7. Flowchart of the GA-MISMO modeling strategy

• Operation. Designing the operators for selection, crossover, and mutation is a major issue in a GA implementation and is always done ad hoc. In each generation, after the fitness values of all the chromosomes in the same population are calculated, the selection operator that chooses chromosomes from the current generation's population for inclusion in the next generation's population is executed. There are many ways for choosing parents [35]. In this study, the roulette method is used for Logistic time series and top percent method is adopted for Mackey-Glass and NN3 time series through preliminary simulation in a trial-error fashion. After selection of parents, the crossover operator should be applied to produce two offspring by exchanging some genetic information between the parents. Crossover occurs during evolution according to the crossover probability. The crossover probability is how often a crossover will be performed. In this study, the probability of crossover is set to 0.9, as shown in Table III; two-point operator is used. This process of selection and reproduction is repeated until the number of offspring becomes equal to the number of eliminated chromosomes; by adding the offspring to population, its size becomes equal to the initial size [34]. The mutation is the third operator, which is applied on the population (parents and their offspring) excluding the best chromosome. A simple mutation operator that selects a few percent of the genes of the population randomly and flips their values from zero to one and vice versa is used in this study [36]. In this study, the probability of mutation is set to 0.02, as shown in Table III. The above process is repeated until a termination conditions are reached. Once the termination conditions are met, the best chromosome $\mathbf{P^*}$ and the corresponding $H$- step-ahead forecast $\{\hat{\varphi}_{N+1}^{*}, \ldots, \varphi_{N+H}^{*}\}$ for chromosome **P\*** are obtained; otherwise, go back to the previous step.

### D. Experimental Procedure

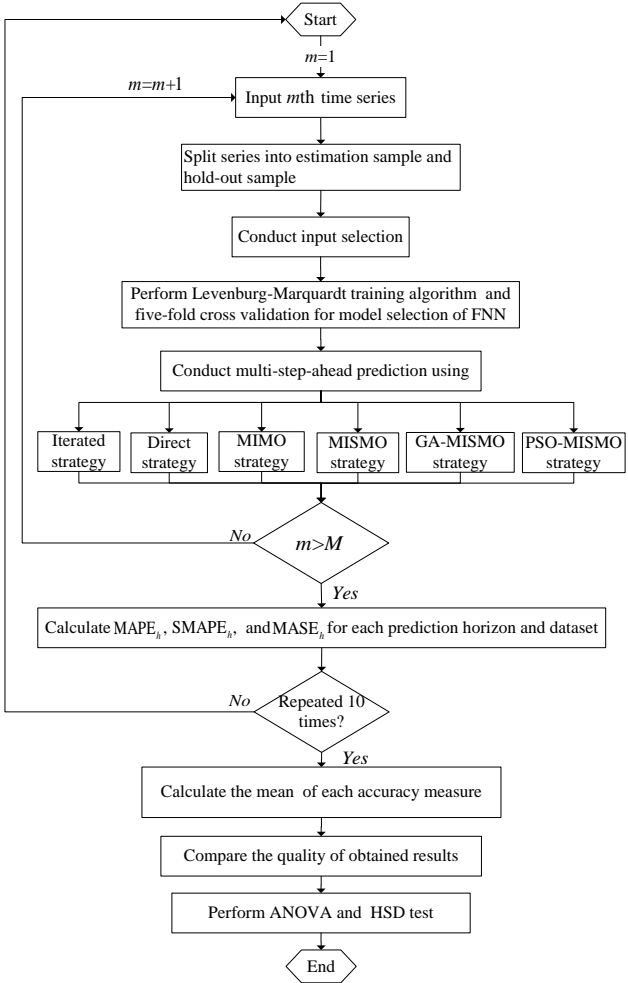

Fig. 8. Experiment procedure

Fig. 8 shows the experimental procedure using the simulated and real time series. Each series is first split into the estimation sample and the hold-out sample. Then, the input selection and model selection for each series are determined using the filter method, the Levenberg and Marquardt algorithm, and fivefold cross-validation with iterated, direct, MIMO, MISMO, GA-MISMO, and PSO-MISMO strategies. Finally the models are tested with the hold-out samples and the $MAPE_h$, $SMAPE_h$ and $MASE_h$ are computed for each prediction horizon $h$ (in this case $h=1,2, \ldots ,18$ ) over datasets (i.e., Logistic, Mackey-Glass, and NN3 datasets). The modeling process for each series is repeated ten times. Upon the termination of this loop, performance of the examined models with selected strategies at each prediction horizon and dataset is judged in terms of the mean, averaged by ten, of the $MAPE_h$, $SMAPE_h$ and $MASE_h$. Analysis of variance (ANOVA) test procedures are used to determine if the means of performance measures are significantly different among the three models for each prediction horizon and datasets. If so, Tukey's honesty significant difference (HSD) tests [37] are then used to further identify the significantly different prediction models in multiple pair-wise comparisons.

## VII. RESULTS AND DISCUSSION

### A. PSO-MISMO vs. the Others in Prediction Accuracy

The prediction performances of all six modeling strategies examined in terms of the three accuracy measures (MAPE, SMAPE, and MASE) and average rank for three datasets are shown in Table IV. The columns labeled as 'Average 1- $h$' show the average accuracy measures over the prediction horizon 1 to $h$ . The last column shows the average ranking for each model over all prediction horizons of the out-of-sample prediction performance. For each column of Table IV, the entry with the smallest value is set in boldface and marked with an asterisk, and the entry with second smallest value is set in boldface type.

The results in Table IV lead to the following conclusions:

• The rankings from best to worst are: PSO-MISMO and GA-MISMO are almost a tie, then MISMO, MIMO, direct, and iterated strategies, regardless of the accuracy measures considered. Thus our findings are robust to the choice of accuracy measures.

• The modeling strategies based on the multiple-output structure (i.e., PSO-MISMO, GA-MISMO, MISMO, and MIMO strategies) outperform those based on the single-output structure (i.e., iterated and direct strategies), which is in agreement with [7].

• Despite the popularity of the iterated strategy in multi-step-ahead prediction literature, it is consistently the worst performing strategy for prediction.

• Comparing the iterated strategy with the direct strategy, the direct strategy is better regardless of the accuracy measures considered, which demonstrates that the accumulation of errors in the case of iterated strategy drastically deteriorates the accuracy of the prediction.

• The MIMO strategy consistently achieves more accurate forecasts than either the iterated or the direct strategies for all prediction horizons. It is conceivable that the reason for the superiority of the MIMO strategy is that it preserves, among the predicted values, the stochastic dependency characterizing the time series.

• The MISMO strategy seems to produce forecasts which are more accurate than those of the MIMO strategy (though only marginally). It is conceivable that the reason for the superiority of the MISMO strategy is that it trades off the property of preserving the stochastic dependency between future values with a greater flexibility of the predictor.

• When comparing the results of two heuristic-based modeling strategies (i.e., PSO-MISMO and GA-MISMO) with those of MISMO strategy, the heuristic-based strategies are generally better, which may indicate that the fixed number of outputs of every sub-model derived from the MISMO strategy tends to prevent the MISMO strategy from considering complex dependencies between prediction steps within different sub models and consequently reduces the prediction accuracy.

TABLE IV
PREDICTION ACCURACY MEASURES FOR HOLD-OUT SAMPLE

| Dataset | Strategy | Prediction horizon( $h$ ) | | | | | | Average 1- $h$ | | | Average Rank |
|---|---|---|---|---|---|---|---|---|---|---|---|
| | | 1 | 2 | 3 | 6 | 12 | 18 | 1-3 | 1-12 | 1-18 | |
| | | **MAPE** | | | | | | | | | |
| | ITER | **47.89** | 39.94 | 39.41 | 43.28 | 35.21 | 51.41 | 42.41 | 43.18 | 46.18 | 5.888 |
| | DIR | 47.93 | 37.18 | 36.27 | 41.71 | 36.14 | 49.22 | 40.46 | 42.95 | 44.87 | 5.000 |
| | MIMO | 47.96 | 38.42 | 37.81 | 40.18 | 34.81 | 47.81 | 41.39 | 42.54 | 42.18 | 4.000 |
| | MISMO | 47.96 | 37.64 | 35.54 | 41.21 | 34.15 | 45.12 | 40.38 | 41.54 | 41.84 | 3.111 |
| | GA-MISMO | **47.88*** | **37.04*** | **35.28** | **39.84** | **31.01*** | **43.01** | **40.06*** | **39.24** | **39.14** | **1.555** |
| | PSO-MISMO | 47.95 | 37.17 | **35.17*** | **38.28*** | 32.73 | **42.18*** | 40.09 | **38.91*** | **38.51*** | **1.444*** |
| | | **SMAPE** | | | | | | | | | |
| | ITER | 31.10 | 33.42 | 36.74 | 39.74 | 45.12 | 39.21 | 33.75 | 39.13 | 40.14 | 5.888 |
| | DIR | **30.97** | 32.02 | 37.18 | 36.18 | 45.85 | 37.44 | 33.39 | 38.67 | 38.79 | 5.055 |
| | MIMO | 31.18 | 32.59 | 36.88 | 34.79 | 44.14 | 35.42 | 33.55 | 38.04 | 38.17 | 3.944 |
| Logistic | MISMO | **30.87*** | **31.98** | 36.31 | 34.82 | 42.74 | 34.18 | 33.05 | 37.87 | 37.49 | 3.166 |
| | GA-MISMO | 31.09 | **31.86*** | **36.14** | 32.97 | **40.81*** | **32.95** | 33.03 | **35.41*** | 35.82 | **1.555** |
| | PSO-MISMO | 31.11 | 32.03 | **35.74*** | **33.58*** | 41.58 | **32.07*** | **32.96*** | 36.13 | **35.18*** | **1.500*** |
| | | **MASE** | | | | | | | | | |
| | ITER | 0.706 | 0.521 | 0.911 | 0.914 | 1.184 | 1.054 | 0.713 | 0.941 | 1.124 | 5.944 |
| | DIR | 0.705 | **0.501** | 0.894 | 0.881 | 0.914 | 0.867 | 0.700 | 0.887 | 0.914 | 4.944 |
| | MIMO | 0.696 | 0.539 | 0.814 | 0.784 | 0.846 | 0.814 | 0.683 | 0.761 | 0.824 | 4.111 |
| | MISMO | **0.691*** | 0.541 | 0.795 | 0.718 | 0.819 | 0.748 | 0.675 | 0.745 | 0.794 | 3.000 |
| | GA-MISMO | 0.705 | 0.505 | **0.731*** | **0.624*** | **0.718*** | 0.642 | 0.647 | **0.687*** | **0.678*** | **1.444*** |
| | PSO-MISMO | **0.695** | **0.499*** | **0.742** | **0.648** | **0.743** | **0.618*** | **0.645*** | **0.702** | **0.704** | 1.555 |
| | | **MAPE** | | | | | | | | | |
| | ITER | 5.15 | 5.26 | 6.34 | 7.15 | 10.18 | 9.46 | 5.58 | 9.01 | 9.22 | 5.944 |
| | DIR | **5.11*** | 4.93 | 5.87 | 6.92 | 8.55 | 8.49 | 5.30 | 7.83 | 8.13 | 5.055 |
| | MIMO | 5.16 | 5.02 | 5.19 | 6.31 | 6.24 | 6.87 | 5.12 | 6.37 | 6.38 | 3.777 |
| | MISMO | **5.13** | **4.88** | 5.23 | 5.87 | 5.97 | 6.05 | 5.08 | 5.89 | 5.64 | 3.222 |
| | GA-MISMO | 5.16 | **4.87*** | **4.61** | **4.87*** | **4.28*** | **4.22*** | **4.88** | **4.78*** | **4.71*** | **1.444*** |
| | PSO-MISMO | 5.15 | 4.91 | **4.50*** | **5.06** | 4.58 | 4.31 | **4.85*** | 4.96 | 4.87 | 1.555 |
| | | **SMAPE** | | | | | | | | | |
| | ITER | **5.28*** | 5.31 | 5.48 | 8.10 | 10.18 | 11.18 | 5.35 | 8.91 | 10.52 | 5.555 |
| | DIR | 5.30 | 5.11 | **4.75** | 7.29 | 8.46 | 9.85 | 5.05 | 7.84 | 9.86 | 5.000 |
| Mackey-Glass | MIMO | **5.29** | 5.25 | 5.62 | 6.57 | 6.87 | 7.46 | 5.38 | 6.05 | 8.05 | 4.055 |
| | MISMO | 5.31 | **5.05*** | 5.32 | 6.18 | 5.94 | 6.28 | 5.22 | 5.87 | 6.91 | 3.111 |
| | GA-MISMO | **5.29** | 5.10 | **4.75** | **5.40** | **4.91** | **4.31*** | 5.04 | **5.16** | 5.19 | **1.611*** |
| | PSO-MISMO | 5.31 | **5.07** | **4.64*** | **5.23*** | **4.72*** | 4.45 | **5.01*** | **5.12*** | **5.03*** | 1.777 |
| | | **MASE** | | | | | | | | | |
| | ITER | 1.785 | 1.683 | 1.628 | 1.935 | 2.148 | 2.184 | 1.698 | 1.924 | 2.015 | 5.722 |
| | DIR | 1.783 | **1.672** | 1.531 | 1.910 | 1.924 | 1.924 | 1.662 | 1.875 | 1.987 | 4.833 |
| | MIMO | **1.769*** | 1.694 | 1.654 | 1.875 | 1.816 | 1.794 | 1.705 | 1.846 | 1.914 | 4.000 |
| | MISMO | **1.775** | 1.681 | 1.615 | 1.849 | 1.748 | 1.649 | 1.690 | 1.789 | 1.841 | 3.055 |
| | GA-MISMO | 1.781 | **1.663*** | **1.522** | **1.746*** | **1.601** | 1.478 | **1.655*** | **1.710** | 1.647 | 1.666 |
| | PSO-MISMO | 1.777 | 1.675 | **1.520*** | **1.759** | **1.588*** | **1.462*** | 1.657 | **1.703*** | **1.638*** | **1.611*** |
| | | **MAPE** | | | | | | | | | |
| | ITER | 14.93 | 13.69 | 17.05 | 19.76 | 18.48 | 29.49 | 15.10 | 22.38 | 23.27 | 5.944 |
| | DIR | **14.87*** | 12.23 | 17.04 | 18.87 | 17.08 | 23.84 | 14.71 | 17.29 | 18.74 | 5.000 |
| | MIMO | **14.88** | 11.68 | 17.09 | 18.01 | 16.07 | 21.54 | 14.55 | 16.41 | 17.57 | 4.055 |
| | MISMO | **14.88** | **11.62** | 17.27 | 17.89 | 15.96 | 21.31 | 14.56 | 15.95 | 16.12 | 2.944 |
| | GA-MISMO | **14.88** | 11.63 | **16.67*** | **17.20** | **15.04*** | 20.34 | **14.39*** | 15.68 | 16.03 | 1.611 |
| | PSO-MISMO | 14.89 | **11.54*** | 16.84 | **17.15*** | 15.15 | **20.22*** | 14.42 | **15.61*** | **15.91*** | **1.444*** |
| | | **SMAPE** | | | | | | | | | |
| | ITER | 15.16 | 13.31 | 14.97 | 20.57 | 17.84 | 20.68 | 14.48 | 19.97 | 21.59 | 5.944 |
| | DIR | **15.10** | 11.84 | 14.34 | 18.37 | 15.67 | 19.97 | 13.76 | 17.75 | 19.84 | 5.000 |
| NN3 | MIMO | 15.12 | 11.17 | 14.22 | 16.37 | 14.83 | 18.34 | 13.50 | 16.58 | 18.34 | 4.055 |
| | MISMO | **15.10** | 10.97 | **13.37** | 15.97 | 14.56 | 18.16 | 13.14 | 16.13 | 17.10 | 2.944 |
| | GA-MISMO | **15.09*** | 10.88 | **13.21*** | 15.86 | **13.43*** | **17.24*** | **13.06*** | **15.61*** | **16.47*** | **1.500*** |
| | PSO-MISMO | 15.11 | **10.68*** | 13.46 | **15.91*** | 13.61 | 17.38 | 13.08 | 15.72 | 16.51 | 1.555 |
| | | **MASE** | | | | | | | | | |
| | ITER | 0.835 | 0.845 | 1.104 | 1.297 | 0.184 | 1.418 | 0.928 | 1.164 | 1.187 | 5.944 |
| | DIR | **0.829*** | 0.859 | 1.197 | 1.168 | 0.108 | 1.387 | 0.962 | 0.109 | 1.046 | 5.055 |
| | MIMO | **0.834** | 0.746 | 1.019 | 1.035 | 0.981 | 1.197 | 0.866 | 0.943 | 0.981 | 3.888 |
| | MISMO | **0.834** | **0.741** | 0.996 | 1.007 | 0.975 | 1.184 | 0.857 | 0.923 | 0.934 | 3.055 |
| | GA-MISMO | 0.835 | **0.731*** | **0.987** | 0.912 | **0.921*** | 1.083 | 0.856 | 0.910 | 0.917 | 1.555 |
| | PSO-MISMO | **0.834** | 0.742 | **0.974*** | **0.907*** | 0.926 | **1.061*** | **0.850*** | **0.903*** | **0.911*** | **1.500*** |

*Note:* For each column of table, the entry with the smallest value is set in boldface and marked with an asterisk, and the entry with second smallest value is set in boldface type.

TABLE V
MULTIPLE COMPARISON RESULTS WITH RANKED STRATEGIES FOR HOLD-OUT SAMPLE

| dataset | Measure | Prediction horizon ($h$) | Rank of strategies | | | | | |
|---|---|---|---|---|---|---|---|---|
| | | | 1 | 2 | 3 | 4 | 5 | 6 |
| Logistic | $MAPE_h$ | 4, 5, 8, 13, 16, 17 | PSO-MISMO < | GA-MISMO <* | MISMO < | MIMO <* | DIR <* | ITER |
| | | 7, 9-11, 15 | GA-MISMO < | PSO-MISMO <* | MISMO < | MIMO <* | DIR <* | ITER |
| | | 18 | PSO-MISMO < | GA-MISMO <* | MISMO <* | MIMO <* | DIR <* | ITER |
| | | 12, 14 | GA-MISMO < | PSO-MISMO <* | MISMO < | MIMO < | ITER < | DIR |
| | | 6 | PSO-MISMO < | GA-MISMO < | MIMO < | MISMO < | DIR <* | ITER |
| | | 2 | GA-MISMO < | PSO-MISMO < | MISMO < | DIR < | MIMO <* | ITER |
| | | 3 | PSO-MISMO < | GA-MISMO < | MISMO < | DIR < | MIMO <* | ITER |
| | $SMAPE_h$ | 6, 11, 14, 15, 18 | PSO-MISMO < | GA-MISMO <* | MISMO < | MIMO <* | DIR <* | ITER |
| | | 5, 8, 13, 16 | GA-MISMO < | PSO-MISMO <* | MISMO < | MIMO <* | DIR <* | ITER |
| | | 7 | PSO-MISMO < | GA-MISMO <* | MISMO <* | MIMO <* | DIR <* | ITER |
| | | 10 | PSO-MISMO < | MIMO < | MISMO < | | DIR <* | ITER |
| | | 12, 17 | GA-MISMO < | PSO-MISMO <* | MISMO <* | MIMO < | ITER < | DIR |
| | | 4, 9 | PSO-MISMO < | GA-MISMO < | MISMO < | DIR < | MIMO <* | ITER |
| | $MASE_h$ | 5, 10, 14, 15 | PSO-MISMO < | GA-MISMO <* | MISMO < | MIMO <* | DIR <* | ITER |
| | | 7, 8, 11, 16, 17 | GA-MISMO < | PSO-MISMO <* | MISMO < | MIMO <* | DIR <* | ITER |
| | | 9, 13 | PSO-MISMO < | GA-MISMO <* | MISMO <* | DIR < | MIMO <* | ITER |
| | | 4, 12 | GA-MISMO < | PSO-MISMO < | MISMO < | MIMO <* | DIR <* | ITER |
| | | 18 | PSO-MISMO < | GA-MISMO <* | MISMO < | MIMO < | DIR < | ITER |
| | | 3 | GA-MISMO < | PSO-MISMO < | MISMO < | MIMO < | DIR <* | ITER |
| | | 6 | PSO-MISMO < | GA-MISMO <* | MISMO <* | MIMO <* | DIR < | ITER |
| Mackey-glass | $MAPE_h$ | 5, 7, 10, 15 | PSO-MISMO < | GA-MISMO <* | MISMO < | MIMO <* | DIR <* | ITER |
| | | 8, 9, 13, 16 | GA-MISMO < | PSO-MISMO <* | MISMO < | MIMO <* | DIR <* | ITER |
| | | 4, 11, 14 | PSO-MISMO < | GA-MISMO <* | MISMO <* | MIMO <* | DIR <* | ITER |
| | | 12, 17 | GA-MISMO < | PSO-MISMO < | MISMO < | MIMO < | ITER < | DIR |
| | | 18 | GA-MISMO < | PSO-MISMO <* | MIMO < | MISMO < | DIR <* | ITER |
| | | 6 | GA-MISMO < | PSO-MISMO < | MIMO < | MISMO <* | DIR < | ITER |
| | | 3 | PSO-MISMO < | GA-MISMO <* | MIMO < | MISMO <* | DIR <* | ITER |
| | $SMAPE_h$ | 6, 9, 11, 13, 15 | PSO-MISMO < | GA-MISMO <* | MISMO < | MIMO <* | DIR <* | ITER |
| | | 4, 7, 8, 14, 16, 17 | GA-MISMO < | PSO-MISMO <* | MISMO < | MIMO <* | DIR <* | ITER |
| | | 12, 10 | PSO-MISMO < | GA-MISMO <* | MISMO <* | MIMO <* | DIR <* | ITER |
| | | 18 | GA-MISMO < | PSO-MISMO <* | MISMO <* | MIMO <* | ITER < | DIR |
| | | 5 | GA-MISMO < | PSO-MISMO < | MISMO < | MIMO <* | ITER < | DIR |
| | | 3 | PSO-MISMO < | GA-MISMO < | DIR < | MISMO <* | ITER <* | MIMO |
| | $MASE_h$ | 7, 10, 11, 14, 16 | PSO-MISMO < | GA-MISMO <* | MISMO < | MIMO <* | DIR <* | ITER |
| | | 5, 8, 13 | GA-MISMO < | PSO-MISMO <* | MISMO < | MIMO <* | DIR <* | ITER |
| | | 18, 15 | PSO-MISMO < | GA-MISMO <* | MISMO <* | MIMO <* | DIR <* | ITER |
| | | 4, 17 | GA-MISMO < | PSO-MISMO < | MISMO < | MIMO < | ITER <* | DIR |
| | | 6, 9 | GA-MISMO < | PSO-MISMO < | MISMO < | MIMO < | DIR <* | ITER |
| | | 3 | PSO-MISMO < | GA-MISMO <* | DIR <* | MISMO < | ITER < | MIMO |
| | | 12 | PSO-MISMO < | GA-MISMO <* | MISMO < | MIMO <* | ITER < | DIR |
| NN3 | $MAPE_h$ | 4, 6, 13,15,18 | PSO-MISMO < | GA-MISMO <* | MISMO < | MIMO <* | DIR <* | ITER |
| | | 5, 8, 10-12 | GA-MISMO < | PSO-MISMO <* | MISMO < | MIMO <* | DIR <* | ITER |
| | | 7, 16, 17 | PSO-MISMO < | GA-MISMO <* | MISMO < | MIMO < | DIR < | ITER |
| | | 9, 14 | GA-MISMO < | PSO-MISMO <* | MISMO < | MIMO <* | DIR <* | ITER |
| | | 2, | PSO-MISMO < | MISMO < | GA-MISMO < | MIMO < | DIR <* | ITER |
| | $SMAPE_h$ | 5, 7-9, 14,17 | PSO-MISMO < | GA-MISMO <* | MISMO < | MIMO <* | DIR <* | ITER |
| | | 4, 10, 11, 12, 16 | GA-MISMO < | PSO-MISMO <* | MISMO < | MIMO <* | DIR <* | ITER |
| | | 6 | PSO-MISMO < | GA-MISMO < | MISMO < | MIMO <* | DIR <* | ITER |
| | | 13 | PSO-MISMO < | MISMO < | GA-MISMO < | MIMO <* | DIR <* | ITER |
| | | 3, 15 | GA-MISMO < | PSO-MISMO <* | MISMO <* | MIMO < | ITER < | DIR |
| | | 18 | GA-MISMO < | PSO-MISMO < | MISMO < | MIMO <* | DIR <* | ITER |
| | | 2 | PSO-MISMO < | GA-MISMO < | MISMO < | MIMO <* | DIR <* | ITER |
| | $MASE_h$ | 5, 7, 8, 11 | PSO-MISMO < | GA-MISMO <* | MISMO < | MIMO <* | DIR <* | ITER |
| | | 10, 13, 14, 16 | GA-MISMO < | PSO-MISMO <* | MISMO < | MIMO <* | DIR <* | ITER |
| | | 9, 15, 18 | PSO-MISMO < | GA-MISMO <* | MISMO < | MIMO < | DIR < | ITER |
| | | 4, 12, 17 | GA-MISMO < | PSO-MISMO <* | MISMO < | MIMO <* | DIR <* | ITER |
| | | 2 | GA-MISMO < | MISMO < | PSO-MISMO < | MIMO <* | DIR <* | ITER |
| | | 6 | PSO-MISMO < | GA-MISMO < | MISMO < | MIMO <* | DIR <* | ITER |
| | | 3 | PSO-MISMO < | GA-MISMO < | MISMO < | MIMO <* | ITER < | DIR |

*Note:* * indicates the mean difference between the two adjacent strategies is significant at the 0.05 level

- As far as the comparison between PSO-MISMO and GA-MISMO is concerned, they are almost a tie and the results are mixing among the prediction measures examined. In terms of MAPE, PSO-MISMO wins in Logistic dataset and NN3 dataset but loses in Mackey-Glass dataset. In terms of SMAPE, PSO-MISMO loses in Mackey-Glass dataset and NN3 dataset but wins in Logistic dataset. In terms of MASE, PSO-MISMO wins in Mackey-Glass dataset and NN3 dataset but loses in Logistic dataset.

Following the experimental procedure presented in Fig. 8, an ANOVA procedure is performed to determine if there exists a statistically significant difference among the six modeling strategies in the hold-out sample for each of the performance measures and the prediction horizon. The results are not included in detail to save space. All ANOVA results are significant at the 0.05 level (with a few exceptions), suggesting that there are significant differences among the six modeling strategies. To further identify the significant difference between any two strategies, the Tukey's HSD test is used to compare all pair-wise differences simultaneously. Tukey's HSD test is a post-hoc test, meaning that Tukey's HSD test should not be performed unless the results of the ANOVA procedure are positive. Table V shows the results of the multiple comparison tests for three datasets. For each accuracy measure and prediction horizon, the strategies are rank ordered from 1 (the best) to 6 (the worst). Several observations can be made from Table V:

- For each accuracy measure and dataset, PSO-MISMO and GA-MISMO significantly outperforms MISMO for the overwhelming majority of prediction horizons.
- Considering the two heuristic-based modeling strategies, we can see that, whatever the dataset used, whatever the accuracy measure considered, and whatever the prediction horizon examined, the difference in prediction performance between PSO-MISMO and GA-MISMO is not significant at the 0.05 level.
- As far as the comparison MISMO vs. MIMO is concerned, the difference in prediction performance is not significant at the 0.05 level, even with a few exceptions
- Generally, the strategies based on the single-output structure perform significantly worse than those based on the multiple-output structure.
- Concerning the current two leading strategies, the direct strategy significantly outperforms the iterated strategy for the majority of prediction horizons.
- The iterated strategy performs the poorest at 95% statistical confidence level in most cases, even with some exceptions.

### B. PSO-MISMO vs. GA-MISMO in Convergence and Computational Time

Up to now, we have compared the proposed PSO-MISMO with other five competitors (i.e., iterated, direct, MIMO, MISMO, and GA-MISMO) in terms of prediction accuracy. In this subsection, the convergence and computational time of the two heuristic-based prediction strategies (i.e., PSO-MISMO and GA-MISMO) are further examined.

It should be noted that although we have used a number of time series for each datasets, i.e., Logistic, Mackey-Glass, and NN3 datasets, general results do not change much within the time series for each datasets. Therefore, to save space, we report only the results for randomly selected three time series from Logistic, Mackey-Glass, and NN3 datasets, respectively. They are the No.1 series from Logistic time series with $\varphi_1 = 0.100$ and length=485, the No.18 series from Mackey-Glass time series with $\varphi_1 = 2$, $\tau = 17$, and length=964, and the No.55 series from NN3 dataset. In addition, in the following experiment, the swarm/population size is fixed to 20 for both PSO-MISMO and GA-MISMO strategies, and generation number is set to 100, as described in Section V. Each strategy is run ten times over each of these time series, in an attempt to eliminate the influence of a lucky initial solution. All the numerical experiments are performed on a personal computer, Inter(R) Core(TM) 2 Duo CPU 2.50 GHz, 1.87-GB memory, and MATLAB environment (Version R2009b).

As was mentioned above, the learning process is repeated 10 times for both PSO-MISMO and GA-MISMO for each time series. Thus, typical convergences of the best and average fitness values as a function of generations for both PSO-MISMO and GA-MISMO on three aforementioned time series are shown in Fig. 9(a)-(c), respectively. As it can be seen, in a typical PSO-MISMO optimization process the best fitness value decreases rapidly and converges at about 48 generations for No.1 Logistic time series, 65 generations for No.18 Mackey-Glass time series, and 59 generations for No. 55 time series in NN3 datasets; whereas for GA-MISMO it takes about 63 generations for No.1 Logistic time series, 77 generations for No.18 Mackey-Glass time series, and 68 generations for No. 55 time series in NN3 datasets. To show how the evolution process is going on for both PSO-MISMO and GA-MISMO, the convergence of the average fitness values are also shown in Fig. 9. Looking at Fig. 9, it is clear that PSO-MISMO seems to perform better than GA-MISMO. Thus, for the present problem the performance of the PSO-MISMO is better than GA-MISMO from an evolutionary point of view.

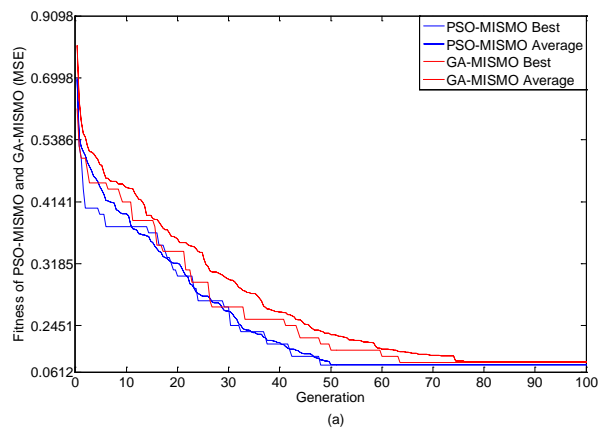

(a)

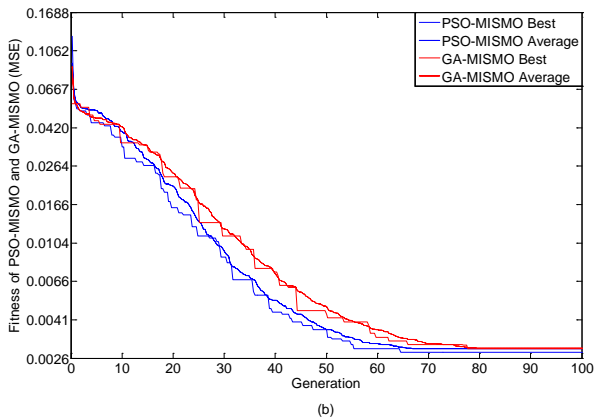

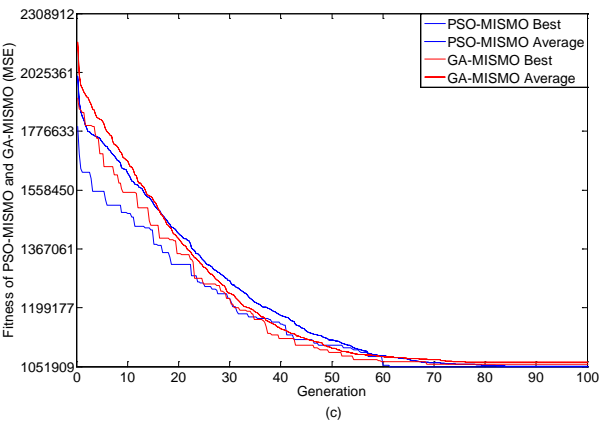

Fig. 9. Convergence of best and average fitness for PSO-MISMO and GA-MISMO strategies. (a) No.1 Logistic time series with $\varphi_1 = 0.100$ and length=485. (b) No.18 Mackey-Glass time series with $\varphi_1 = 2$, $\tau = 17$, and length=964.(c) No.55 time series in NN3 dataset.

The second test performed is designed to measure the average computational time needed per time series from all the datasets within these two strategies. The result is shown in Table VI, from which we can see that the PSO-MISMO needs generally less time than GA-MISMO.

TABLE VI
COMPUTATIONAL TIME FOR PSO-MISMO AND GA-MISMO

| Dataset | Strategy | CPU time (M) |
|---|---|---|
| Logistic dataset | PSO-MISMO | 24.514 |
| | GA-MISMO | 38.421 |
| Mackey-Glass dataset | PSO-MISMO | 30.824 |
| | GA-MISMO | 41.725 |
| NN3 dataset | PSO-MISMO | 11.854 |
| | GA-MISMO | 19.627 |

Generally speaking, thus, from the above experiments, we can draw the following three main conclusions. (1) Two heuristic-based modeling strategies (i.e., PSO-MISMO and GA-MISMO) are capable of obtaining high-quality multi-step-ahead time series forecasts related to the four well-established strategies. (2) The performance of the PSO-MISMO is better than that of GA-MISMO from an evolutionary point of view. (3) PSO-MISMO showed to be faster than GA-MISMO in terms of average running time.

## VIII. CONCLUSIONS

Multi-step-ahead time series prediction has usually proved to be an intractable task due to the growing amount of uncertainties arising from various sources. For instance, an accumulation of errors and lack of information make multi-step-ahead prediction more difficult. Thus, modeling strategies for multi-step-ahead prediction are major research topics with significant practical implications. The contribution of this study is an extension to a well-established MISMO modeling strategy by incorporating a heuristic based on binary particle swarm optimization into the MISMO modeling process to self-adaptively determine the number of sub-models with varying prediction horizons, and conduct a large scale comparative study with neural networks for validation. Quantitative and comprehensive assessments are performed with the simulated and real time series on the basis of the prediction accuracy, convergence, and computational time. Experimental results and comparisons demonstrate the superiority of the proposed PSO-MISMO modeling strategy for multi-step-ahead time series prediction.

## REFERENCES

[1] A. F. Atiya, S. M. El-Shoura, S. I. Shaheen, and M. S. El-Sherif, "A comparison between neural-network forecasting techniques-case study: river flow forecasting," *Neural Networks, IEEE Transactions on*, vol. 10, pp. 402-409, 1999.

[2] G. Chevillon, "DIRECT MULTI - STEP ESTIMATION AND FORECASTING," *J. Econ. Surveys*, vol. 21, pp. 746-785, 2007.

[3] A. Sorjamaa, J. Hao, N. Reyhani, Y. Ji, and A. Lendasse, "Methodology for long-term prediction of time series," *Neurocomputing*, vol. 70, pp. 2861-2869, 2007.

[4] A. Sorjamaa and A. Lendasse, "Time series prediction using DirRec strategy," presented at the European Symposium on Artificial Neural Networks Bruges, 2006.

[5] G. Bontempi, "Long term time series prediction with multi-input multi-output local learning," in *Proceedings of the 2nd European Symposium on Time Series Prediction (TSP), ESTSP08*, Helsinki, Finland, 2008, pp. 145-154.

[6] S. Ben Taieb, G. Bontempi, A. Sorjamaa, and A. Lendasse, "Long-term prediction of time series by combining direct and MIMO strategies," in *Proceedings of the 2009 IEEE International Joint Conference on Neural Networks*, Atlanta, U.S.A., 2009, pp. 3054-3061.

[7] S. Ben Taieb, A. Sorjamaa, and G. Bontempi, "Multiple-output modeling for multi-step-ahead time series forecasting," *Neurocomputing*, vol. 73, pp. 1950-1957, 2010.

[8] S. Ben Taieb, G. Bontempi, A. F. Atiya, and A. Sorjamaa, "A review and comparison of strategies for multi-step ahead time series forecasting based on the NN5 forecasting competition," *Expert Syst. Appl.*, 2012.

[9] H. S. Hippert, C. E. Pedreira, and R. C. Souza, "Neural networks for short-term load forecasting: A review and evaluation," *Power Systems, IEEE Transactions on*, vol. 16, pp. 44-55, 2001.

[10] A. Sharma, "Seasonal to interannual rainfall probabilistic forecasts for improved water supply management: Part 1—A strategy for system predictor identification," *J Hydrol*, vol. 239, pp. 232-239, 2000.

[11] G. P. Zhang and D. M. Kline, "Quarterly time-series forecasting with neural networks," *Neural Networks, IEEE Transactions on*, vol. 18, pp. 1800-1814, 2007.

[12] J. Kennedy and R. Eberhart, "Particle swarm optimization," in *Proceedings of the IEEE International Conference on Neural Networks*, 1995, pp. 1942-1948 vol. 4.

[13] C. F. Juang, "A hybrid of genetic algorithm and particle swarm optimization for recurrent network design," *Systems, Man, and Cybernetics, Part B: Cybernetics, IEEE Transactions on*, vol. 34, pp. 997-1006, 2004.

[14] Y. P. Chen, W. C. Peng, and M. C. Jian, "Particle swarm optimization with recombination and dynamic linkage discovery," *Systems, Man, and Cybernetics, Part B: Cybernetics, IEEE Transactions on,* vol. 37, pp. 1460-1470, 2007.

[15] B. Liu, L. Wang, and Y. H. Jin, "An effective PSO-based memetic algorithm for flow shop scheduling," *Systems, Man, and Cybernetics, Part B: Cybernetics, IEEE Transactions on,* vol. 37, pp. 18-27, 2007.

[16] R. Xu, J. Xu, and D. Wunsch, "A Comparison Study of Validity Indices on Swarm-Intelligence-Based Clustering," *Systems, Man, and Cybernetics, Part B: Cybernetics, IEEE Transactions on,* vol. 42, pp. 1243-1256, 2012.

[17] T. Nguyen, S. Yang, and C. Li, "A self-learning particle swarm optimizer for global optimization problems," 2012.

[18] J. Kennedy and R. C. Eberhart, "A discrete binary version of the particle swarm algorithm," in *Systems, Man, and Cybernetics, 1997. Computational Cybernetics and Simulation., 1997 IEEE International Conference on,* Orlando, FL, 1997, pp. 4104-4108 vol. 5.

[19] K. De Jong, "Parameter setting in EAs: a 30 year perspective," in *Parameter Setting in Evolutionary Algorithms,* ed: Springer, 2007, pp. 1-18.

[20] I. C. Trelea, "The particle swarm optimization: convergence analysis and parameter selection," *Inform Process Lett,* vol. 85, pp. 317-325, 2003.

[21] J. F. Kennedy, J. Kennedy, and R. C. Eberhart, *Swarm intelligence*: Morgan Kaufmann Pub, 2001.

[22] Y. Shi and R. C. Eberhart, "Empirical study of particle swarm optimization," in *Evolutionary Computation, 1999. CEC 99. Proceedings of the 1999 Congress on,* 1999.

[23] B. Liu, L. Wang, and Y.-H. Jin, "An effective PSO-based memetic algorithm for flow shop scheduling," *Systems, Man, and Cybernetics, Part B: Cybernetics, IEEE Transactions on,* vol. 37, pp. 18-27, 2007.

[24] G. Rubio, H. Pomares, I. Rojas, and L. J. Herrera, "A heuristic method for parameter selection in LS-SVM: Application to time series prediction," *Int J Forecasting,* vol. 27, pp. 725-739, 2011.

[25] L.-C. Chang, P.-A. Chen, and F.-J. Chang, "Reinforced Two-Step-Ahead Weight Adjustment Technique for Online Training of Recurrent Neural Networks," *Neural Networks and Learning Systems, IEEE Transactions on,* vol. 23, pp. 1269-1278, 2012.

[26] Y. Liu and X. Yao, "Simultaneous training of negatively correlated neural networks in an ensemble," *Systems, Man, and Cybernetics, Part B: Cybernetics, IEEE Transactions on,* vol. 29, pp. 716-725, 1999.

[27] R. M. May, "Simple mathematical models with very complicated dynamics," *Nature,* vol. 261, pp. 459-467, 1976.

[28] M. C. Mackey and L. Glass, "Oscillation and chaos in physiological control systems," *Science,* vol. 197, pp. 287-289, 1977.

[29] D. E. Goldberg, "Genetic algorithms in search, optimization, and machine learning," *Reading, MA: Addison-Wesley,* 1989.

[30] K. M. Sim, "BLGAN: Bayesian learning and genetic algorithm for supporting negotiation with incomplete information," *Systems, Man, and Cybernetics, Part B: Cybernetics, IEEE Transactions on,* vol. 39, pp. 198-211, 2009.

[31] L. Davis, "Handbook of genetic algorithms," *New York: Van Nostrand Reinhold,* 1991.

[32] D. Golmohammadi, R. C. Creese, H. Valian, and J. Kolassa, "Supplier selection based on a neural network model using genetic algorithm," *Neural Networks, IEEE Transactions on,* vol. 20, pp. 1504-1519, 2009.

[33] D. E. Goldberg and K. Deb, "A comparative analysis of selection schemes used in genetic algorithms," *Urbana,* vol. 51, pp. 61801-2996, 1991.

[34] M. V. Baghmisheh, K. Madani, and A. Navarbaf, "A discrete shuffled frog optimization algorithm," *Artif Intell Rev,* vol. 36, pp. 267-284, 2011.

[35] R. L. Haupt and S. E. Haupt, *Practical genetic algorithms*: Wiley-Interscience, 2004.

[36] E. Elbeltagi, T. Hegazy, and D. Grierson, "Comparison among five evolutionary-based optimization algorithms," *Adv Eng Inform,* vol. 19, pp. 43-53, 2005.

[37] F. Ramsay and D. Schaefer, *The Statistical Sleuth*: Duxbury, Boston, Mass, 1996.

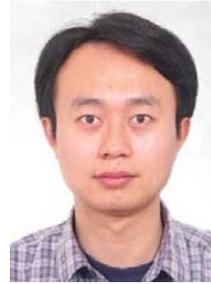

**Yukun Bao** received his B.Sc., M.Sc., and Ph.D in management science and engineering from Huazhong University of Science and Technology, P.R.China, in 1996, 1999 and 2002 respectively.

Currently, he is an associate professor at the department of management science and information systems, school of management, Huazhong University of Science and Technology, P.R.China. He has been the principal investigator for two research projects funded by Natural Science Foundation of China and has served as a referee of paper review for several IEEE journals and international journals and a PC member for several international academic conferences. His research interests are time series modeling and forecasting, business intelligence and data mining.

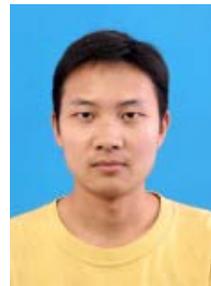

**Tao Xiong** received his B.Sc. and M.Sc. in management science and engineering from Huazhong University of Science and Technology, P.R.China, in 2008 and 2010 respectively.

Currently, he is working toward the Ph.D. degree in management science and engineering, Huazhong University of Science and Technology, China. His research interests include multi-step-ahead time series forecasting, interval data analysis, computational intelligence.

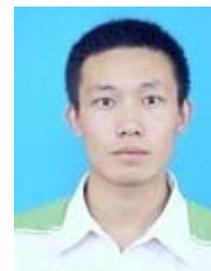

**Zhongyi Hu** received B.Sc. degree in information system in 2009 from Harbin Institute of Technology at Weihai, P.R.China. He received his M.Sc. degree in management science and engineering in 2011 from Huazhong University of Science and Technology, P.R.China.

Currently, he is working toward the Ph.D. degree in management science and engineering, Huazhong University of Science and Technology, P.R.China. His research interests include support vector machines, swarm intelligence, memetic algorithms, and time series forecasting.